\pdfoutput=1

\documentclass[11pt]{article}

\usepackage[]{acl}

\usepackage{times}
\usepackage{latexsym}

\usepackage[T1]{fontenc}

\usepackage[utf8]{inputenc}

\usepackage{microtype}
\usepackage{graphicx}
\usepackage{booktabs}
\usepackage{tabularx}
\usepackage{multirow}
\usepackage{amsmath}
\usepackage{gb4e}

\newcommand{\f}{\mkern-2mu f\mkern-1mu}

\usepackage{color}
\definecolor{Red}{RGB}{163, 0, 0}
\definecolor{Blue}{RGB}{0, 0, 163}
\newcommand{\blue}[1]{\textcolor{Blue}{#1}}
\newcommand{\red}[1]{\textcolor{Red}{#1}}

\title{When a sentence does not introduce a discourse entity,\\ Transformer-based models still sometimes refer to it}

\author{
 Sebastian Schuster \\
 Center for Data Science\\
 Department of Linguistics\\
 New York University\\
 schuster@nyu.edu\\ \And
 Tal Linzen \\
  Center for Data Science\\
 Department of Linguistics\\
 New York University\\
 linzen@nyu.edu
}

\date{}

\begin{document}
\maketitle
\begin{abstract}
Understanding longer narratives or participating in conversations requires tracking of discourse entities that have been mentioned. Indefinite noun phrases (NPs), such as \textit{a dog}, frequently introduce discourse entities but this behavior is modulated by sentential operators such as negation. For example, \textit{a dog} in \textit{Arthur doesn't own a dog} does not introduce a discourse entity due to the presence of negation. In this work, we adapt the psycholinguistic assessment of language models paradigm to higher-level linguistic phenomena and introduce an English evaluation suite that targets the knowledge of the interactions between sentential operators and indefinite NPs. We use this evaluation suite for a fine-grained investigation of the entity tracking abilities of the Transformer-based models GPT-2 and GPT-3. We find that while the models are to a certain extent sensitive to the interactions we investigate, they are all challenged by the presence of multiple NPs and their behavior is not systematic, which suggests that even models at the scale of GPT-3 do not fully acquire basic entity tracking abilities. 
\end{abstract}

\section{Introduction}

In order to understand longer narratives or to participate in conversations, humans and natural language understanding systems have to keep track of the entities that have been mentioned in the discourse. For example, when someone tells you that \textit{Arthur owns a dog}, they have introduced the entity of a person named \textit{Arthur} and the entity of a dog owned by Arthur into the discourse. Once entities have been introduced to the discourse, it is natural to refer back to them with noun phrases or pronouns to elaborate further, e.g., by saying \textit{It has a red collar} to elaborate on the dog. 

While no fully-specified account exists of how humans achieve this feat, many existing theories are based on the idea that humans maintain mental files \cite[e.g.,][]{heim1982semantics,murez2016mental}, i.e., explicit memory representations for each entity that encode all properties of an entity and its relation to other entities. When engaging in a conversation or reading a longer narrative, humans then update these representations as they encounter new entities or new information about existing entities.

Large pre-trained language models (LMs) such as GPT-2 \cite{radford2019language} and GPT-3 \cite{brown2020language}, which in recent years have become the dominant foundation for natural language understanding and generation tasks, lack explicit representations of discourse entities. It remains largely an open question to what extent LMs can match human behavior in tracking discourse entities.

The most extensive investigation of this phenomenon has been through evaluations with the LAMBADA dataset \cite{paperno-etal-2016-lambada}. LAMBADA consists of a cloze task for which an LM has to predict the last word of naturalistic passages extracted from a corpus. 
Solving this task requires keeping track of longer contexts, and making a correct guess frequently requires keeping track of the entities mentioned in the passage. 

While datasets such as LAMBADA are an invaluable resource for monitoring high-level progress of LMs in their ability to track discourse entities, such datasets lack the granularity to determine for which contexts LMs can and cannot properly track discourse entities. In this work, we draw inspiration from recent targeted evaluation suites geared at lower linguistic levels \cite[e.g.,][]{marvin-linzen-2018-targeted,hu-etal-2020-systematic}, and introduce a targeted evaluation suite for tracking of discourse entities in English. In particular, we focus on the interactions between different sentential operators and embedding verbs and indefinite noun phrases (see, e.g., \citealt{karttunen1976discourse} and Section~\ref{sec:background}); for example, we evaluate whether a language model correctly infers that because a sentence with a negation, such as \textit{Arthur \textbf{doesn't} own a dog}, does not introduce a discourse entity for a dog, further elaborations about such a non-existent dog should be pragmatically odd, and, as such, their probability should be low compared to matched controls.

To evaluate to what extent language models are sensitive to these interactions, we adapt the psycholinguistic assessment of language models paradigm \cite{futrell-etal-2019-neural} for discourse entity tracking and discuss the methodological challenges that arise with using this paradigm for discourse phenomena. We introduce two expert-created evaluation suites and use them to evaluate GPT-2 and GPT-3 models. We find that while all the models we evaluated show some sensitivity to preceding context, they lack systematicity and are challenged when contexts contain multiple noun phrases.\footnote{Our evaluation suites along with the results from human experiments and all code for model evaluation is available at \url{https://github.com/sebschu/discourse-entity-lm}.}

\section{Related Work}

The majority of systematic evaluations of autoregressive and masked language models so far have focused on the level of syntax, targeting abilities such as subject-verb agreement \citep[e.g.,][]{linzen-etal-2016-assessing,marvin-linzen-2018-targeted,gulordava-etal-2018-colorless,hu-etal-2020-systematic}, anaphora agreement and binding constraints \citep[e.g.,][]{marvin-linzen-2018-targeted,futrell-etal-2019-neural,warstadt-etal-2020-blimp-benchmark, hu-etal-2020-closer}, or filler-gap dependencies \citep[e.g.,][]{wilcox-etal-2018-rnn,chowdhury-zamparelli-2018-rnn,da-costa-chaves-2020-assessing}.
At higher linguistic levels, \newcite{ettinger-2020-bert} compared BERT's \cite{devlin-etal-2019-bert} behavior on sentences with negation to data from neurolinguistic experiments with humans; \newcite{pandia-ettinger-2021-sorting} investigated to what extent pre-trained language models can extract relevant information from the preceding context, both in the presence and in the absence of distractors; and \newcite{pandia-etal-2021-pragmatic} investigated whether language models can predict connectives (e.g., \textit{but}) for two given sentences.

More closely related to our work, in the domain of discourse and reference,  \newcite{upadhye-etal-2020-predicting} investigated whether GPT-2 and Transformer-XL \cite{dai-etal-2019-transformer} exhibit similar referential biases in their continuations as humans, i.e., they asked whether models provided with a sentence with two referents are biased similarly as humans when choosing the referent for the next sentence. \newcite{kim-etal-2019-probing} used an acceptability  judgment task to determine whether contextual LMs correctly distinguish between definite and indefinite noun phrases.

\newcite{sorodoc-etal-2020-probing} and \newcite{tenney-etal-2019-bert} used probing methods to investigate whether representations of LSTM- and Transfomer-based models contain information about coreference, which also provides some indication of entity tracking abilities. Further, \newcite{clark-etal-2019-bert} investigated to what extent attention weights of BERT indicate coreference. These studies found that all evaluated representations contain some information about coreference but not consistently for all entities. 


\section{Background}
\label{sec:background}

English indefinite noun phrases (NPs) of the form \textit{a NOUN} interact with the context in complex ways \cite[see, e.g.,][for more extensive discussions of this phenomenon]{karttunen1976discourse,Webber1979,heim1982semantics}. In affirmative statements, the indefinite NP generally introduces a new entity to the discourse. 
However, several sentential operators and clause-embedding verbs modulate this behavior. For example, consider the following contrast between an affirmative and a negated sentence, where \# indicates a pragmatically odd continuation:

\begin{exe}
\ex \begin{xlist}
\ex \label{ex:affirmative1} Arthur owns a dog and it follows him everywhere he goes.
\ex \label{ex:negation1}  Arthur doesn't own a dog and \# it follows him everywhere he goes.
\end{xlist}
\end{exe}
While in the affirmative sentence, the indefinite NP introduces a novel discourse entity, the negation in (\ref{ex:negation1}) prevents the NP from introducing a new entity. In (\ref{ex:negation1}), it is therefore pragmatically odd to refer back to \textit{a dog} with the pronoun \textit{it}.

The implicative \textit{manage to} and the negative implicative \textit{fail to} in (\ref{ex:managed-failed1}a-b) give rise to a similar contrast: The NP under \textit{manage to} introduces a discourse entity, the NP under \textit{fail to} does not.
\begin{exe}
\ex \label{ex:managed-failed1} \begin{xlist}
\ex  \label{ex:managed1} Sue managed to write a book. It was a real page-turner.
\ex  \label{ex:failed1}  Sue failed to write a book. \# It was a real page-turner.
\end{xlist}
\end{exe}

Indefinite NPs embedded under the factive \textit{know} and the non-factive \textit{doubt} introduce and fail to introduce a discourse entity, respectively:
\begin{exe}
\ex \label{ex:know-doubt1} \begin{xlist}
\ex \label{ex:know1} I know that Michael baked a cake. It was delicious. 
\ex \label{ex:doubt1} I doubt that Michael baked a cake. \# It was delicious.
\end{xlist}
\end{exe}

Lastly, modals such as \textit{want} also block the introduction of a discourse entity, as shown in (\ref{ex:affirmative-modal1}):
\begin{exe}
\ex \label{ex:affirmative-modal1} \begin{xlist}
\ex \label{ex:affirmative2} Mary got a pet rat and it is very loud at night. 
\ex \label{ex:modal1} Mary wants to get a pet rat and \# it is very loud at night.
\end{xlist}
\end{exe}

While these patterns generally hold, there are exceptions to these rules. For example, in the first sentence in (\ref{ex:specific1}), which could be paraphrased as (\ref{ex:specific-para}), the indefinite scopes over the negation and therefore it is okay to refer back to the mistake.

\begin{exe}
\ex \label{ex:specific1} Mary didn't find a (specific) mistake. It was in the footnote. 
\ex \label{ex:specific-para} There was a (specific) mistake which Mary did not find. It was in the footnote. 
\end{exe} 	 

However, without additional context, listeners generally do not infer these so-called specific interpretations of sentences with an indefinite NP, so like \newcite{karttunen1976discourse}, we will largely ignore these cases throughout the remainder of this paper.

\section{Experiments}
To what extent are GPT-2 and GPT-3 sensitive to the contrasts that we presented in Section~\ref{sec:background}? To investigate this question, we adapted the methodology commonly used for syntactic evaluation of language models \cite[e.g.,][]{futrell-etal-2019-neural} and created minimal pairs of contexts that differ in whether they introduce a discourse entity or not. In Experiment~1, we focus on contexts with a single indefinite NP, and in Experiment~2, we focus on sentences with multiple indefinite NPs.  

\subsection{Experiment 1}
\label{sec:exp-1}

\paragraph{Methodology} If a language model is sensitive to contexts that differ in whether a discourse entity is introduced or not, we expect the probability of continuations that refer back to the noun phrase in the previous context to be higher when a discourse entity has been introduced than when it has not. Thus, if we have a pair of sentences, such as 
\begin{exe}
 \ex \begin{xlist}
 \ex C$_{{re\f}}$: John owns a dog. 
 \ex C$_{{nonre\f}}$: John doesn't own a dog.
 \end{xlist}
\end{exe}
and a referential continuation,\footnote{The psycholinguistic assessment of language models paradigm  generally focuses on the probability of individual words rather than sequences to evaluate syntactic phenomena. However, considering that the coreference of \textit{it} (or other referential expressions) is modulated by an entire sentence or clause (see the contrast between (\ref{ex:ref-cont}) and (\ref{ex:non-ref-cont}), which both contain the pronoun \textit{it}), we compare probabilities of sequences.} such as 
\begin{exe}
\ex \label{ex:ref-cont} $R$: It has a red collar.
\end{exe}
then we expect that
$$P(R \mid C_{{re\f}}) > P(R \mid C_{{nonre\f}}).$$

However, directly comparing these probabilities may be problematic given that $P(X \mid C_{{re\f}})$ and $P(X \mid C_{{nonre\f}})$ are different probability distributions. In theory it could be, for example, that $P(X \mid C_{{re\f}})$ assigns a very high probability to exactly one continuation and therefore its entropy is lower than the entropy of $P(X \mid C_{{nonre\f}})$. In this case, it could be that the inequality above does not hold despite the fact that continuations that refer back to the noun phrase in the previous context are ranked higher in the affirmative than in the negated case. To overcome this issue, we make use of a non-referential control continuation, such as N:

\begin{exe}
 \ex \label{ex:non-ref-cont} N: It is not a big deal.
\end{exe}
This continuation no longer refers back to a noun phrase and is therefore a valid continuation for both contexts. Instead of using the inequality above, we thus compare the relative probabilities of the referential and the control continuations:
\begin{align} \label{eq:rel-prob}
& \frac{P(R \mid C_{{re\f}})}{P(R \mid C_{{re\f}}) + P(N \mid C_{{re\f}})} \\ \nonumber > \quad 
&\frac{P(R \mid C_{{nonre\f}})}{P(R \mid C_{{nonre\f}}) + P(N \mid C_{{nonre\f}})}
\end{align}
These relative probabilities are less sensitive to the entropy of the distribution: If there is a highly likely continuation (that is neither the referential nor the control continuation) for one context but not the other, the model should still assign relatively less probability mass to the referential continuation compared to the control continuation. 

\paragraph{Models} We evaluate two autoregressive language models,\footnote{We selected these autoregressive models instead of masked language models (MLMs) such as BERT \cite{devlin-etal-2019-bert} because they are more frequently used to generate texts, and discourse abilities such as entity tracking tend to play a more crucial role in generating text than in classification or span extraction tasks for which MLMs are more frequently used.} GPT-2 and GPT-3. GPT-2 models were trained on the WebText corpus which contains an estimated 8 billion tokens; GPT-3 models were trained on about 500 billion tokens. For GPT-2, we evaluate models of four different sizes (GPT-2: 117M parameters, GPT-2 M: 345M, GPT-2 L: 762M, GPT-2 XL: 1.5B)  that are available through the HuggingFace Transformers library \cite{wolf-etal-2020-transformers}. For GPT-3, we evaluate the largest available model (``davinci'') through the OpenAI API which is assumed to have about 175B parameters.\footnote{The model size of GPT-3 is not publicly available but the EleutherAI project estimated the model size by comparing the performance of the models available through the API to published results: \url{https://blog.eleuther.ai/gpt3-model-sizes/}.}

\begin{table*}[]
    \centering
    \resizebox{\textwidth}{!}{
\begin{tabular}{@{}llll@{}}
\toprule
  \multicolumn{1}{l}{Contrast}                                         & \multicolumn{1}{c}{Contexts}                          & \multicolumn{1}{c}{Referential continuation}                                  & \multicolumn{1}{c}{Non-referential continuation}              \\ \midrule
\multirow{2}{*}{affirmative-negation} & Michael baked a cake              & \multirow{2}{*}{and it was the best thing at the picnic.} & \multirow{2}{*}{and it's not a big deal.} \\
                                      & Michael didn't bake a cake        &                                                           &                                           \\  \midrule
\multirow{2}{*}{affirmative-modal}    & Michael baked a cake              & \multirow{2}{*}{and it was the best thing at the picnic.} & \multirow{2}{*}{and it's not a big deal.} \\
                                      & Michael wants to bake a cake     &                                                           &                                           \\ \midrule
\multirow{2}{*}{know-doubt}           & I know that Michael baked a cake. & \multirow{2}{*}{It was the best thing at the picnic.}     & \multirow{2}{*}{It's not a big deal.}     \\
                                      & I doubt that Michael baked a cake. &                                                           &                                           \\ \midrule
\multirow{2}{*}{managed-failed}       & Michael managed to bake a cake.   & \multirow{2}{*}{It was the best thing at the picnic.}     & \multirow{2}{*}{It's not a big deal.}     \\
                                      & Michael failed to bake a cake.    &                                                           &                                          
\end{tabular}
}
\caption{\label{tbl:stimuli-exp1}Example contexts and continuations for one base context in Experiment~1.}
\end{table*}

\paragraph {Materials} We manually constructed an evaluation set of 16 base contexts and plausible continuations. Each base context contains different nouns and verbs to minimize lexical effects. From these 16 contexts, we constructed four contrasts for each context, as shown in Table~\ref{tbl:stimuli-exp1}, which in total yielded 64 items. We chose to manually construct contexts as opposed to generating sentences from a grammar to guarantee semantic and pragmatic well-formedness of contexts and continuations. 

\paragraph{Human evaluation} As we mentioned in Section~\ref{sec:background}, the referential continuations are not necessarily pragmatically odd if the indefinite noun phrase in the context is interpreted as a specific noun phrase. We therefore conducted an online experiment on Prolific to verify that native English speakers disprefer the referential continuations when no discourse entity is introduced, as follows. After two practice items, each participant performed two trials with sentences from the evaluation set. On each trial, participants saw a context along with a referential and a non-referential continuation, and they were asked to indicate their preferred continuation by selecting the continuation that ``makes more sense'' given the context. For each context, we collected preference judgments from 10 participants. The experiment took on average about 2 minutes to complete and participants received \$0.45 in compensation ($\sim$\$14/hr).

\begin{figure*}
    \centering
    \includegraphics[width=\textwidth]{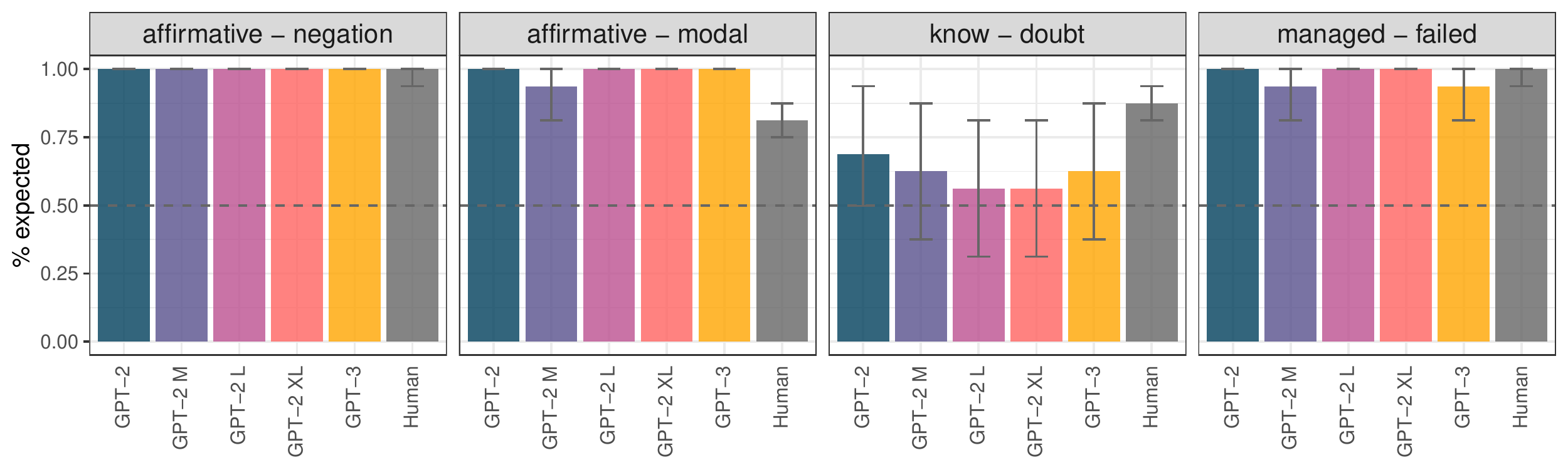}
    \caption{Results from Experiment 1. Each bar indicates the proportion of items for which the relative probability of the referential continuation (RRP) is higher for the context that introduces a discourse entity than for the context that does not, i.e., the expected pattern. Dashed lines indicate chance performance levels, and error bars indicate bootstrapped 95\% confidence intervals.}
    \label{fig:results-exp1}
\end{figure*} 

\paragraph{Results and discussion}
Figure~\ref{fig:results-exp1} shows the proportion of items for which the relative probability of the referential continuation (RRP) is higher for the context that introduces a discourse entity (DEC) than for the context that does not (NDEC), i.e., the proportion of items for which Eq.~\ref{eq:rel-prob} holds. For three of the four contrasts (\textit{affirmative-negation}, \textit{affirmative-modal}, \textit{managed-failed}) GPT-2 and GPT-3 models exhibited the expected pattern for almost all items in our evaluation set. For the \textit{know-doubt} contrast, however, all models performed approximately at chance, suggesting that the models are not sensitive to this contrast.

Figure~\ref{fig:results-exp1} also shows the results of the human experiment. Participants preferred the referential continuation more following the DECs than following the NDECs for all items of the \textit{affirmative-negation} and \textit{managed-failed} contrasts. Further, for these two contrasts, participants overwhelmingly selected the referential continuation for the DECs and the non-referential continuation for the NDECs. This result confirms that the stimuli bring about the theoretically expected contrast in humans.

For the \textit{affirmative-modal} and the \textit{know-doubt} contrasts, the results from human participants are less clear-cut. Overall, participants also preferred the referential continuation more in the DECs than in the NDECs. However, for several items, the opposite was the case and the referential continuation was preferred as much or even more in the NDECs than in the DECs. Moreover, unlike in the other two contrasts, participants selected the  referential continuation in the NDECs at a high rate.\footnote{For contexts with modals, some participants commented that they selected the referential continuation because they assumed that the past tense of the continuation was a grammatical mistake. That is, if the tense had been different, the continuation would have been sensible. For example, for the context \textit{Michael wants to bake a cake} the continuation \textit{and it \textbf{will be} the best thing at the picnic} is acceptable and differs from the continuation that was presented in the experiment, \textit{and it was the best thing at the picnic}, only in its tense.

For contexts with \textit{doubt}, participants frequently seemed to interpret the referential continuation as a reason for the doubt. For example, for the context \textit{I doubt that Carla got a pet rat.}, participants frequently chose the referential continuation \textit{It is very noisy at night.},  presumably because they considered that the rat being noisy made it unlikely that Carla would have got it.}

Considering that the results from the human experiment are not predicted by Karttunen's theory, the model results from the \textit{affirmative-modal} and the \textit{know-doubt} contrast should also be interpreted with caution. However, while the lower proportion of expected relative probabilities in the \textit{know-doubt} condition may suggest that the models are behaving similarly to humans, this is not the case. If one considers the results on an item-by-item basis, they differ from the human results and there is a lot of variability across models such that the five models agree only on less than 33\% of items.

In summary, GPT-2 and GPT-3 overall behaved similarly to humans and generally favored the referential continuation more when the preceding sentence introduced a discourse entity. This behavior could be due to at least the following two reasons. It could be that the models indeed correctly combine the sentential operators with the indefinite noun phrase and therefore assign a higher probability to a referential continuation in the DECs. However, it could also be that this result is due to more spurious correlations; for example, it could be that the model learned that clauses with operators such as negation, modals, or negative implicatives are often followed by clauses with a non-referential \textit{it}. In the second experiment, we tease apart these two explanations and further try to overcome the issues with the stimuli that we observed for the \textit{affirmative-modal} and \textit{know-doubt} contrasts.

\subsection{Experiment 2}

\begin{table*}[]
    \resizebox{\textwidth}{!}{ 
   \begin{tabular}{@{}lll@{}} \toprule
\multicolumn{1}{c}{Context}                                & \multicolumn{1}{c}{Coreferential continuations} & \multicolumn{1}{c}{Non-coreferential continuations} \\ \midrule
Mary \blue{found a shirt} at the store but she \red{didn't find a hat}. & The \blue{shirt}/\red{\#hat} is blue.                           & The \red{hat}/\blue{\#shirt} that she tried on didn't fit.             \\
Mary \blue{found a hat} at the store but she \red{didn't find a shirt}. & The \blue{hat}/\red{\#shirt} is blue.                             & The \red{shirt}/\blue{\#hat} that she tried on didn't fit.           \\
Mary \red{didn't find a shirt} at the store but she \blue{found a hat}. & The \blue{hat}/\red{\#shirt} is blue.                             & The \red{shirt}/\blue{\#hat} that she tried on didn't fit.           \\
Mary \red{didn't find a hat} at the store but she \blue{found a shirt}. & The \blue{shirt}/\red{\#hat} is blue.                           & The \red{hat}/\blue{\#shirt} that she tried on didn't fit.    \\ \bottomrule        
\end{tabular}}
    \caption{Example contexts and continuations for the \textit{affirmative-negation} contrast for one base context.}
    \label{tbl:stimuli-exp2}
\end{table*}

\paragraph{Materials and method}
We again constructed 16 base contexts that are similar to the ones used in Experiment~1. However, in this experiment, each context contains two indefinite noun phrases with different nouns that are embedded under two different sentential operators. For example, for the \textit{affirmative-negation} contrast, one of the NPs is embedded under negation, such as \textit{a cat} in (\ref{ex:exp2-context}).
\begin{exe}
\ex \label{ex:exp2-context} John owns a dog but he doesn't own a cat.
\end{exe}

\begin{figure*}
    \centering
    \includegraphics[width=\textwidth]{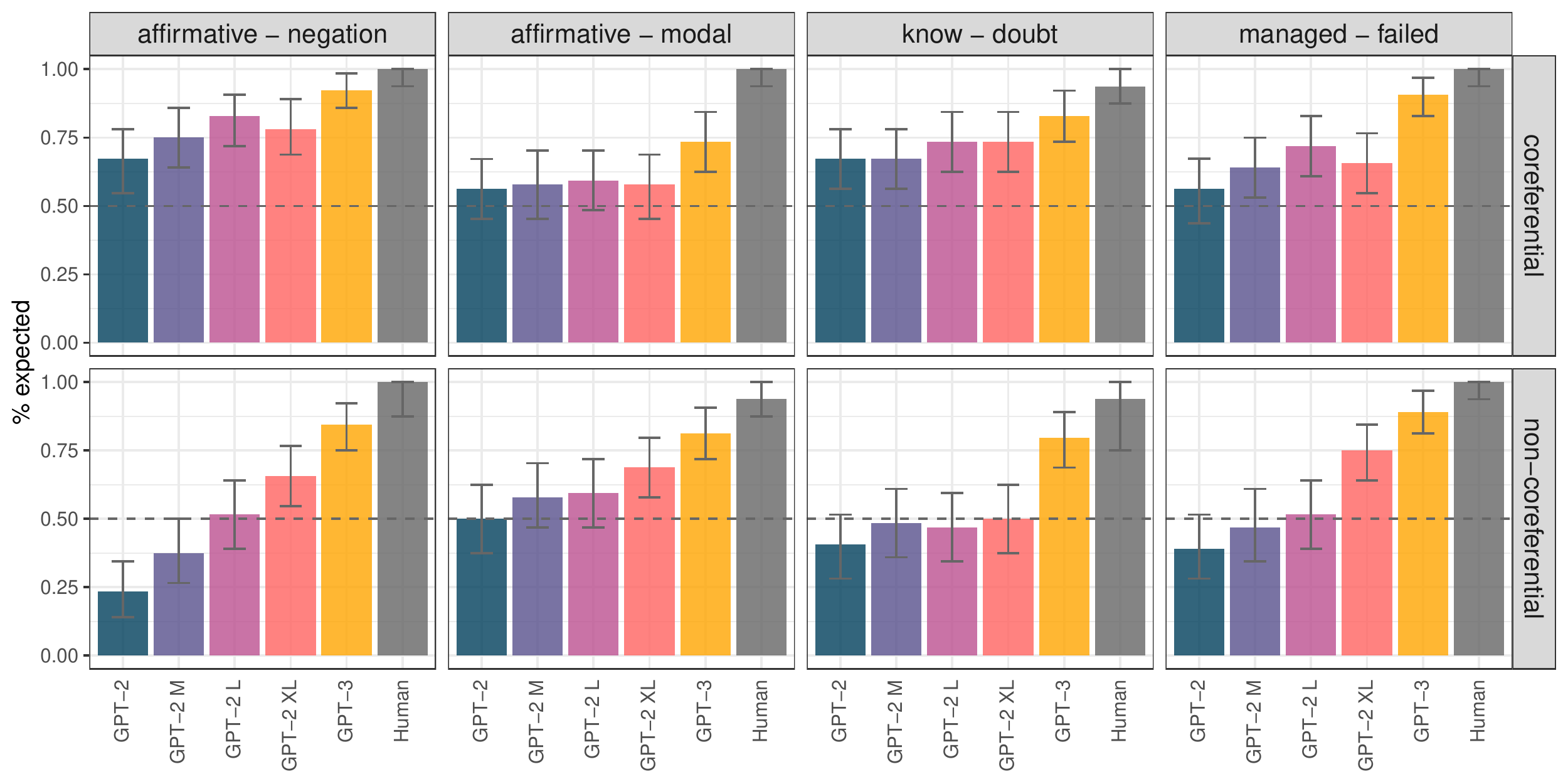}
    \caption{Results from Experiment 2. Dashed lines indicate chance performance levels.}
    \label{fig:results-exp2}
\end{figure*} 

In such a context, it is natural to continue with a sentence that refers back to the dog, whereas it is unnatural to refer back to a cat. We therefore compared the models' probability of a sentence that refers back to an entity that has been introduced in the context (\ref{ex:exp2-cont-ref-fel}) to a sentence that refers to an entity that has not been introduced (\ref{ex:exp2-cont-ref-infel}).
\begin{exe}
\ex 
\begin{xlist}
\ex \label{ex:exp2-cont-ref-fel} The dog follows him wherever he goes.
\ex \label{ex:exp2-cont-ref-infel}\# The cat follows him wherever he goes.
\end{xlist}
\end{exe}

On top of these coreferential continuations, we also constructed non-coreferential continuations for contexts such as (\ref{ex:exp2-context}). These continuations contain one of the nouns present in the context but do not refer back to entities in the previous context. For the non-coreferential continuations, models should assign a higher probability to the continuation with a noun for which no discourse entity had been introduced in the context.
\begin{exe}
\ex 
\begin{xlist}
\ex \label{ex:exp2-cont-nonref-fel} The cat that he liked had been adopted by someone else.
\ex \label{ex:exp2-cont-nonref-infel}\# The dog that he liked had been adopted by someone else.
\end{xlist}
\end{exe}

For each of the four contrasts and each base context, we constructed four contexts that crossed the order of the sentential operators and the order of the two nouns used in a context, resulting in 4 contexts per base context and contrast. For each base context, we further constructed two coreferential continuations (one for each noun) and two non-coreferential continuations (one for each noun). In total, this yielded 512 items. Table~\ref{tbl:stimuli-exp2} shows all the contexts and continuations for one base context for the \textit{affirmative-negation} contrast.

Compared to the methods and materials in Experiment~1, this setup has several advantages. First, given that we are comparing two continuations for a fixed context, both continuations come from the same probability distribution and therefore we no longer need a generic control continuation. Second, it is less likely that models can make use of spurious correlations since each context contains two types of sentential operators and, for example, a heuristic of associating negation with non-referential \textit{it} would no longer lead to the expected behavior. Third, given that all continuations are on topic (as opposed to the generic control condition in Experiment~1), humans likely show more consistency in their preferences. Lastly, given that this design allows us to construct stimuli with exactly the same tokens in different orders, we can also assess the systematicity of the model behavior.

We again verified the theoretically predicted preferences in a human experiment.\footnote{For practical reasons, we included two items from this experiment in the first human experiment. To rule out interference between similar items, no two items within the same experimental list were derived from the same base context.}


\begin{figure*}
    \centering
    \includegraphics[width=\textwidth]{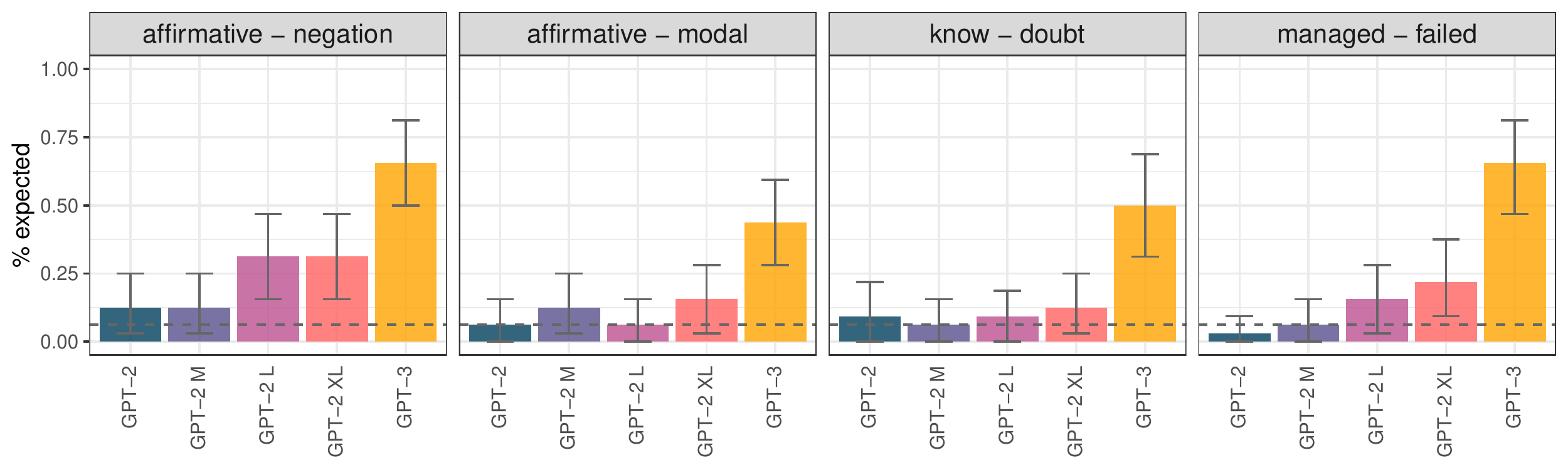}
    \caption{Systematicity of model behavior in Experiment 2. An item counts as correct if all four orders of noun phrases and sentential operators (e.g., \textit{X owns a A but doesn't own a B}; \textit{X owns a B but doesn't own a A}; \textit{X doesn't own a A but owns a B}; and \textit{X doesn't own a B but owns a A})  lead to the correct result. The dashed line indicates chance performance and the error bars indicate bootstrapped 95\% confidence intervals.}
    \label{fig:results-exp2-systematicity}
\end{figure*} 

\paragraph{Results and discussion}
Figure~\ref{fig:results-exp2} shows the accuracy of the model and human experiments for the coreferential and non-coreferential continuations. As this figure shows, humans exhibited the theoretically expected behavior for all contrasts for almost all items and chose the coreferential continuation with the noun for which an entity had been introduced in the context, and chose the non-coreferential continuation for the noun for which no entity had been introduced. This suggests that the materials do not exhibit the same shortcomings as in Experiment~1, and that comparisons of models to human behavior are valid for all four contrasts.

If we turn to the model results, there is more variability in performance across models and contrasts. For the coreferential continuations, all models except the smallest GPT-2 model performed above chance for three of the four contrasts. For the \textit{affirmative-modal} contrast, however, only GPT-3 performed significantly above chance. Moreover, all GPT-2 models perform worse for the non-coreferential continuations.

More generally, unlike humans, all models in this experiment performed below ceiling, which suggests that while models exhibit a tendency to choose the right continuation, they do not reliably do so. Further, model size does have an impact on the performance on this task: The smallest GPT-2 model performed consistently worst, and GPT-3 performed consistently best. This dependence on model size is particularly pronounced in the non-coreferential condition: While GPT-3 consistently performed above chance in all contrasts, most smaller models either performed at chance or in some cases, such as the smallest GPT-2 for the items in the \textit{affirmative-negation} contrast, had a bias to select the non-coreferential continuation with the noun that introduced a discourse entity in the context. The lower performance for the non-coreferential continuations is not surprising given that for these examples, a model not only has to correctly infer which noun phrase introduces a discourse entity but additionally that the noun phrase in the continuation does not refer back to anything in the preceding context.

\paragraph{Systematicity} As mentioned above, this experimental design also allows us to assess how sensitive the behavior of the different models is to the different orders of sentential operators and nouns in the context. Figure~\ref{fig:results-exp2-systematicity} shows the proportion of items for which the model exhibited the expected behavior for all four possible orders. Overall, the performance of all models according to this stricter criterion is much lower than the simple by-item measure highlighting that even the predictions by GPT-3 are sensitive to the exact combination and order of sentential operators and nouns. However, there once again is a clear trend that larger models behave more systematically than smaller ones, suggesting that larger models and models trained on more data learn more stable generalizations. This trend is in part driven by smaller models being less sensitive to the preceding context: The two smallest GPT-2 models assigned the highest probability to the continuation with one of the two nouns independent of the combination of sentential operators and nouns in the context in 52.3\% and 43.8\% of the cases, respectively. That is, for all four contexts, as shown for one example in Table~\ref{tbl:stimuli-exp2}, the smallest GPT-2 model assigned a higher probability to  the same continuation independent of which noun phrase introduced a discourse entity more than half of the time. GPT-3, on the other hand, only exhibited this behavior for 7\% of the items.

In summary, the results from Experiment~2 suggest that GPT-2 and GPT-3 are less reliable in determining whether an NP introduces a discourse entity when multiple NPs are present. This is in particular true for smaller GPT-2 models but if one considers systematicity, the predictions of GPT-3 are also sensitive to minor changes in the context.

\section{Likely Continuations}

One drawback of the methodology of the previous two experiments is that we considered only one specific expected and one specific unexpected continuation for each item. Thus, if both the expected and the unexpected continuations are very unlikely according to the LM, we may see poor performance on this task while at the same time, it would be very unlikely that either of the generations is ever sampled from the model. In that case, the evaluations in Experiment~2 may underestimate the models' abilities \cite{newman-etal-2021-refining} and the results may not be very relevant for practical purposes for which one uses an LM to generate texts. For this reason, we also performed a manual analysis of randomly sampled generations \cite{aina-linzen-2021-language} from the two largest LMs, GPT-2 XL and GPT-3.

\paragraph{Materials and method}

We used the contexts from Experiment~2 as prompts for the two LMs and for each context, we sampled a sentence beginning with \textit{the}.\footnote{As compared to just using the context as a prompt, constraining the continuation to start with \textit{the} led to considerably more continuations with noun phrases referring back to a noun phrase in the context while still putting very few constraints on the overall continuation.} 
For GPT-2 XL, we sampled the continuations using top-$40$ sampling as in \newcite{radford2019language}. For GPT-3, we used the default temperature sampling with a temperature of $0.7$.

A graduate student in linguistics who was blind to the purpose of this study then annotated each of the continuations for whether it contained referring expressions that likely referred back to a noun phrase in the context as well as which noun phrase(s) (the discourse entity introducing and/or the non-discourse entity introducing NP) were referred back to. To illustrate this, consider the following two generations by GPT-3:
\begin{exe}
\ex \label{ex:annotation-exs} \begin{xlist}
\ex {Carolyn didn't write a card to her parents but she wrote them a letter. The} \textbf{letter was long and filled with many details about the cruise.}
\ex {Chris managed to knit a hat but failed to knit a bag. The} \textbf{bag is not stuffed.}
\end{xlist}
\end{exe}
In (\ref{ex:annotation-exs}a), \textit{the letter} refers back to an entity introduced in the context, whereas in (\ref{ex:annotation-exs}b), \textit{the bag} refers back to the NP that does not introduce a discourse entity. If a language model is able to correctly combine sentential operators with indefinite noun phrases, we expect many continuations as in (\ref{ex:annotation-exs}a) and no continuations as in (\ref{ex:annotation-exs}b).

\paragraph{Results and discussion}

\begin{table}[]
    \centering
    \small
     \begin{tabular}{l c c}
        \toprule
        Model & DE & NDE \\
        \midrule
        GPT-2 XL & 43.8 & 22.3 \\
        GPT-3 & 52.3 & 21.1 \\
        \bottomrule
    \end{tabular}
    \caption{Percentage of expressions in model generations that refer back to noun phrases that introduce (DE) or do not introduce a discourse entity (NDE).}
    \label{tbl:continuation-results}
\end{table}

Table~\ref{tbl:continuation-results} shows the percentages of expressions in model generations that refer back to noun phrases in the prompt. These results confirm the findings from Experiment~2: Both GPT-2 XL and GPT-3 are to some extent sensitive to the interactions between sentential operators and indefinite NPs as indicated by the higher proportion of expressions referring back to NPs that introduce discourse entities (DE) as compared to referring back to NPs that do not (NDE). At the same time, however, both models produced more than 20\% of continuations with expressions that refer back to an NP that did not introduce an entity, which shows that the results from Experiment~2 also apply to likely generations by LMs.

\section{General Discussion}

In his seminal work in 1976, \citeauthor{karttunen1976discourse} introduced several challenges for natural language understanding systems that aim to track which entities are introduced in a larger discourse. In this work, we evaluated to what extent we made progress on these challenges in the past decades. In two sets of experiments, we found that Transformer-based models are to some extent sensitive to different interactions between sentential operators and indefinite noun phrases. At the same time, however, we found in Experiment~2 that models lack systematicity in their behavior, which suggests that models do not combine indefinite noun phrases and sentential operators as humans do. Further, the analysis of likely continuations showed that this behavior can also be observed in high probability generations.

\paragraph{Learnability of meaning} On the one hand, these results provide direct evidence for shortcomings of language models with respect to tracking entities. On the other hand, more broadly, these results also provide interesting data points with regard to the recent debate on whether language models could theoretically mimic human language understanding. \newcite{bender-koller-2020-climbing} recently presented several thought experiments and argued that since LMs are only trained on form and do not have access to meaning or intentions, they can never exhibit human-like language understanding  \cite[see also][for a more formal discussion of this claim]{merrill-etal-2021-provable}. Given that we evaluated the largest available GPT-3 model and still found that the model behavior is inconsistent despite its enormous amount of parameters and training data, our results suggest that at least current language model architectures indeed struggle with human-like understanding. Interestingly though, while the thought experiments by \newcite{bender-koller-2020-climbing} focus on lack of world knowledge due to the lack of grounding of language models, our results suggest that additionally, language models fail at learning the meaning of more abstract words such as negation markers or embedding verbs. This is also in line with recent results, which showed that smaller models fail to learn the meaning of negation and discourse connectives. \cite{ettinger-2020-bert,pandia-etal-2021-pragmatic}. Lastly, the fact that GPT-2 and GPT-3 have been exposed to orders of magnitude more language data than human learners are and still do not fully succeed at tracking discourse entities underscores that there are differences between how humans and LMs learn. 

\paragraph{NLG evaluation} We further believe that evaluation suites targeting discourse phenomena, such as the ones presented here, can serve a complementary role to natural language generation (NLG) benchmarks \cite[e.g.,][]{gehrmann-etal-2021-gem} and human studies for evaluating NLG systems. This seems particularly relevant considering that \newcite{clark-etal-2021-thats} recently found that untrained crowdworkers, who serve as participants in the majority of human evaluation studies, cannot distinguish between stories written by humans and stories generated by GPT-3. Our experiments, however,  show that there is a considerable gap between humans and GPT-3 for basic discourse phenomena, and therefore targeted evaluation suites should be an important measure for tracking progress of NLG models. 

\paragraph{Comparison to probing results} Recently, \newcite{li-etal-2021-implicit} developed a probe for investigating whether LM representations provide information about the state of entities at various stages in a larger discourse. This probing method--like the ones presented in this work--also aims to assess entity tracking abilities of pre-trained language models. They considered two sequence-to-sequence models, T5 and BERT, and found that representations from both models can be decoded into entity states with high accuracy. This task may seem more complex than the one used in the experiments above, and T5 and BERT are considerably smaller models than GPT-3, so prima facie, it may be surprising that their results suggest superior discourse abilities than our results. However, there are two important differences in methodology that likely explain this discrepancy. First, the probing classifier was trained on data that was similar to the evaluation data and this setup therefore provided a lot of supervision. Second, the datasets used by \newcite{li-etal-2021-implicit} were obtained through crowdsourcing or a generation engine and were not constructed as systematically as ours. For these reasons, the probing classifier may have learned spurious correlations between the training and test splits, and the high accuracy on the task may have only in part been driven by entity tracking abilities of LMs.

\paragraph{Potential solutions} Considering the still modest performance of GPT-3, it seems unlikely that training models on even more data is going to lead to human-like discourse entity processing by language models. Instead, we consider the following modifications to models to likely lead to more systematic entity tracking. First, there have been some successes in augmenting language models with explicit entity memory representations \cite[e.g.,][]{weston2014memory,sukhbaatar2015endtoend,rashkin-etal-2020-plotmachines,cheng2020attending}, and likely such architectural changes could also help in the contexts that we evaluated in this work. Second, considering that the models seem to struggle to learn the meaning of sentential operators, it may be necessary to provide additional supervision, for example using treebanks annotated with meaning representations, such as the Groningen Meaning Bank \cite{Bos2017GMB}. Relatedly, models may also benefit from more grounded learning scenarios. Humans likely differentiate between \textit{Arthur owns a dog} and \textit{Arthur doesn't own a dog} because only in the former case, \textit{a dog} refers to something in the real world and if a model was immersed in more grounded scenarios it would likely be able to infer this difference. 

We hope that our evaluation suite will be a valuable resource for assessing progress of future models such as the ones sketched above, and that it will help pave the way for improved discourse entity processing in NLU systems.


\section*{Ethics Statement}

\paragraph{Risks, limitations, and intended use} We consider the main risk of this work that the evaluation suite may be used to make overstating claims about model abilities in the future. In particular, should future models achieve very high or even perfect accuracy on the evaluation suite, then such results may be seen as evidence of human-like abilities of discourse entity processing. We therefore want to emphasize that achieving high accuracy on this evaluation suite is a necessary but not necessarily sufficient requirement for a model to exhibit human-like entity tracking abilities. 

Further, it seems likely that models fine-tuned on similar examples would perform a lot better on this evaluation suite, and therefore researchers should only use this dataset for out-of-domain evaluations in which the model has not been trained on similar examples.

Finally, we only evaluated models trained on English data in this work and it is conceivable that entity tracking abilities of models trained on other languages differ from the results reported here. 

\paragraph{Human subject experiments} As we mentioned in Section~\ref{sec:exp-1}, we recruited crowdworkers from Prolific to validate the experimental stimuli. Participants were based in the US and on average received compensation of about \$14/hour, which is almost twice the minimum wage in most states in the US. The experiment has been pre-approved by the New York University IRB, and there were no risks associated with participation.

\section*{Acknowledgments}

We thank the members of the NYU Computation and
Psycholinguistics Lab and the NYU Semantics Group, and the reviewers for their thoughtful feedback. We also thank Alicia Chatten for doing the annotations of the model generations. This material is based upon work supported by the National Science Foundation under Grant \#2030859 to the Computing Research Association for the CIFellows Project. Any opinions, findings, and conclusions or recommendations expressed in this material are those of the authors and do not necessarily reflect the views of the National Science Foundation nor the Computing Research Association.

\bibliography{paper.bib}

\appendix

\section{Human experiment details}

\begin{figure*}
    \centering
    \includegraphics[width=\textwidth]{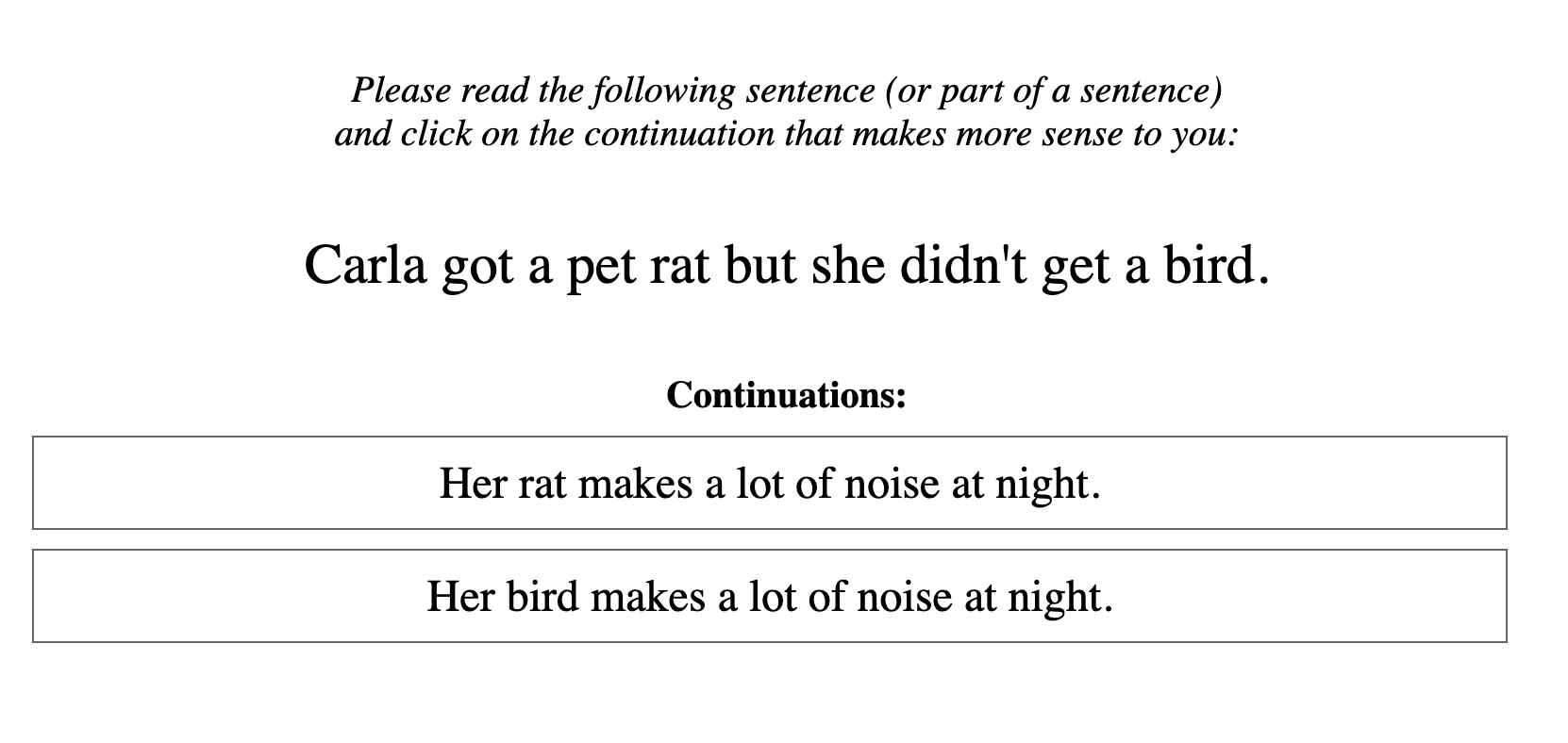}
    \caption{Example trial of human experiment.}
    \label{fig:example-trial}
\end{figure*}

Participants completed two practice trials to get familiarized with the task, followed by four critical trials with two filler trials randomly interspersed. Figure~\ref{fig:example-trial} shows an example trial. Participation was limited to people living in the US whose native language is English.

\section{Model experiment details}

For the experiments with GPT-2, we used the LM-Scorer library\footnote{\url{https://github.com/simonepri/lm-scorer/}} and ran the experiments on a node with a 3.7Ghz CPU and 32GB of RAM. In total, all evaluations required approximately 8h of CPU time. For the experiments with GPT-3, we used the offical OpenAI API.\footnote{\url{https://beta.openai.com}} For all experiments, we compared raw, untransformed probabilities, i.e., the temperature parameter was set to 0.

\end{document}